\documentclass[journal]{IEEEtran}

\usepackage[cmex10]{amsmath}
\usepackage{amssymb}
\usepackage{amsthm,mathrsfs,amsfonts,dsfont} 
\usepackage{algorithm}
\usepackage{algorithmic}
\usepackage{array}
\usepackage{fixltx2e}
\usepackage{graphicx}
\graphicspath{{figures/}}
\usepackage{booktabs}
\usepackage{textcomp}
\usepackage{bm}

\newcommand{\Tamb}{T_{\mathrm{a}}}
\newcommand{\Twater}{T_{\mathrm{in}}}

\newcommand{\dayahead}{24}
\newcommand{\balancing}{34}
\newcommand{\lab}{15}
\newcommand{\liters}{120 }
\newcommand{\simrun}{100 }
\newcommand{\layers}{50 }

\newcommand{\Rea}{{\mathbb{R}}}
\newcommand{\Nat}{{\mathbb{N}}}
\newcommand{\Exp}{{\bf \mathbf{E}\/}}

\newcommand{\etal}{\textit{et al. }}

\newcommand{\uphk}{u_k^{\mathrm{ph}}}

\newcommand{\uphl}{u_l^{\mathrm{ph}}}

\newcommand{\soc}{x_{\mathrm{soc}}}

\newcommand{\socmax}{\bar{x}_{\mathrm{soc}}}
\newcommand{\socmin}{\underline{x}_{\mathrm{soc}}}
\newcommand{\probw}{p_{\mathcal{W}}(\cdot|x)}
\begin{document}
\title{Reinforcement Learning  Applied to an Electric Water Heater: From Theory to Practice}

\author{F. Ruelens, B. J. Claessens, S. Quaiyum, B. De Schutter, R. Babu\v{s}ka, and R. Belmans        
\thanks{
F. Ruelens and R. Belmans  are with the Department of Electrical Engineering, KU~Leuven/EnergyVille, Belgium (frederik.ruelens@esat.kuleuven.be).}
%. 
\thanks{
B.~J. Claessens is with the Energy Department of Vito/EnergyVille, Belgium (bert.claessens@vito.be).}
\thanks{
S. Quaiyum is with the Department of Electrical Engineering,  Uppsala University, Sweden.}
\thanks{
B. De Schutter  and R. Babu\v{s}ka are with the Delft Center for Systems and Control, Delft University of Technology, The Netherlands.}}

\markboth{Transaction on Smart Grid}{}
\maketitle

\begin{abstract}
Electric water heaters have the ability to store energy in their water buffer without impacting the comfort of the end user.
This feature makes them a prime candidate for residential demand response. 
However, the stochastic and non-linear dynamics of electric water heaters, makes it challenging to harness their flexibility. %  for demand response.
Driven by this challenge, this paper formulates the underlying sequential decision-making problem as a Markov decision process and uses techniques from reinforcement learning.
Specifically, we apply an auto-encoder network to find a compact feature representation of the sensor measurements, which helps to mitigate the curse of dimensionality.
A well-known batch reinforcement learning technique, fitted Q-iteration, is used to find a control policy, given this feature representation.
In a simulation-based experiment using an electric water heater with 50 temperature sensors, the proposed method was able to achieve good policies much faster than when using the full state information.
%Compared to a default thermostat controller, the approach successfully reduced the cost of energy consumption.
In a lab experiment, we apply fitted Q-iteration to an electric water heater with eight temperature sensors. 
Further reducing the state vector did not improve the results of fitted Q-iteration.
The  results of the lab experiment, spanning 40 days, indicate that compared to a thermostat controller, the presented approach  was able to reduce the total cost of energy  consumption of the electric water heater by \lab$\%$.
\end{abstract}

\begin{IEEEkeywords}
Auto-encoder network, demand response, electric water heater, fitted Q-iteration, reinforcement learning.
\end{IEEEkeywords}

%% figuurtjes instellen
\newif\iflarge
\newif\ifack
\newif\ifapp

\largefalse
\acktrue
\appfalse

% \fi
\IEEEpeerreviewmaketitle

\section{Introduction}
\label{sec:Introduction}
\IEEEPARstart{T}{he} share of renewable energy sources  is expected to reach $25\%$ of the global power generation  portfolio by 2020~\cite{outlook2013}.
The intermittent and stochastic nature of most renewable energy sources, however, makes it challenging to integrate these sources into the power grid.
%In addition to having a limited controllability, most renewable energy sources,such as wind and solar power, are variable and stochastic, which makes them hard to predict.
%As a result of their limited predictability, keeping the supply and demand of electricity in balance will require new solutions.
Successful integration of these sources  requires flexibility on the demand side through demand response programs.
Demand response programs enable end users with flexible loads to adapt their consumption profile in response to an external signal.
A prominent example of flexible loads are electric water heaters with a hot water storage tank~\cite{hastings1980ten,SMARTBOILER}.
These loads have  the ability to store energy in their water buffer without impacting the comfort level of the end user.
In addition to having significant flexibility, electric water heaters can consume about 2 MWh per year for a household with a daily hot water demand of 100 liters~\cite{energygov}. 
As a result, electric water heaters are a prime candidate for residential demand response programs.
%Several papers in the smart grid literature have investigated the use of electric water heaters for demand response. %~\cite{PDE,koch2011modeling,RuelensBRLCluster,diao2012electric}.
Previously, the flexibility offered by electric water heaters has been used for frequency control~\cite{diao2012electric}, local 
voltage control~\cite{Sandero}, and energy arbitrage~\cite{koch2011modeling,Mathieu}.
Amongst others, two prominent control paradigms in the demand response literature on electric water heaters are model-based approaches and reinforcement learning.
%The first popular control paradigm are rule-based controllers, such as fuzzy logic~\cite{6787100} and droop-based mechanisms~\cite{Sandero,diao2012electric}.
%In general, they are relatively simple to implement and require limited computational processing power, since they require no optimization or forecasting.

Perhaps the most researched control paradigm  applied to demand response are model-based approaches, such as Model Predictive Control (MPC)~\cite{koch2011modeling,sossan2013scheduling,Mathieu}.
Most MPC strategies use a gray-box model, based on general expert knowledge of the underlying system dynamics, requiring a system identification step.
Given this mathematical model, an optimal control action can be found by solving a receding horizon problem~\cite{camacho2004model}. 
%The fact that MPC can exploit expert knowledge and use state-of-the art solvers makes them popular control strategy for cluster control or building control.
In general, the implementation of MPC consists of several critical steps, namely, selecting accurate models, estimating
the model parameters, estimating the state of the system, and forecasting of the exogenous variables. 
All these steps make MPC an expensive technique, the cost of which needs to be balanced out by the possible financial gains~\cite{challengesMPC}.
Moreover, possible model errors resulting from an inaccurate model or forecast, can effect the stability of the MPC controller~\cite{ma2012model,Maasoumy}.

%The third control paradigm is to use Reinforcement Learning (RL) techniques~\cite{sutton1998reinforcement}.
In contrast to MPC, Reinforcement Learning (RL) techniques~\cite{sutton1998reinforcement} do not require expert knowledge and consider their environment as a black-box.
RL techniques enable an agent to learn a control policy by interacting with its environment, without the need to use modeling and system identification techniques.
%A comparison of MPC and a well-established RL technique, fitted Q-iteration, for a power system problem can be found in~\cite{ernst2009reinforcement}.
In~\cite{ernst2009reinforcement},~Ernst \etal state that the trade-off in applying MPC and RL mainly depends on the quality of the expert knowledge about the system dynamics that could be exploited in the context of system identification.
In most residential demand response applications, however, expert knowledge about the system dynamics or future disturbances  might be unavailable or might be too expensive to obtain relative to the expected financial gain.
%For example different residential users will exhibit different behavior patterns and  their flexible load will exhibit different models and model parameters.
%As a result, developing a MPC approach can become to expensive for a large scale demand response roll-out.
In this context, RL techniques are an excellent candidate to build a general purpose agent that can be applied to any demand response application.

\begin{figure}[t]
\centerline{\includegraphics[width=0.38\columnwidth]{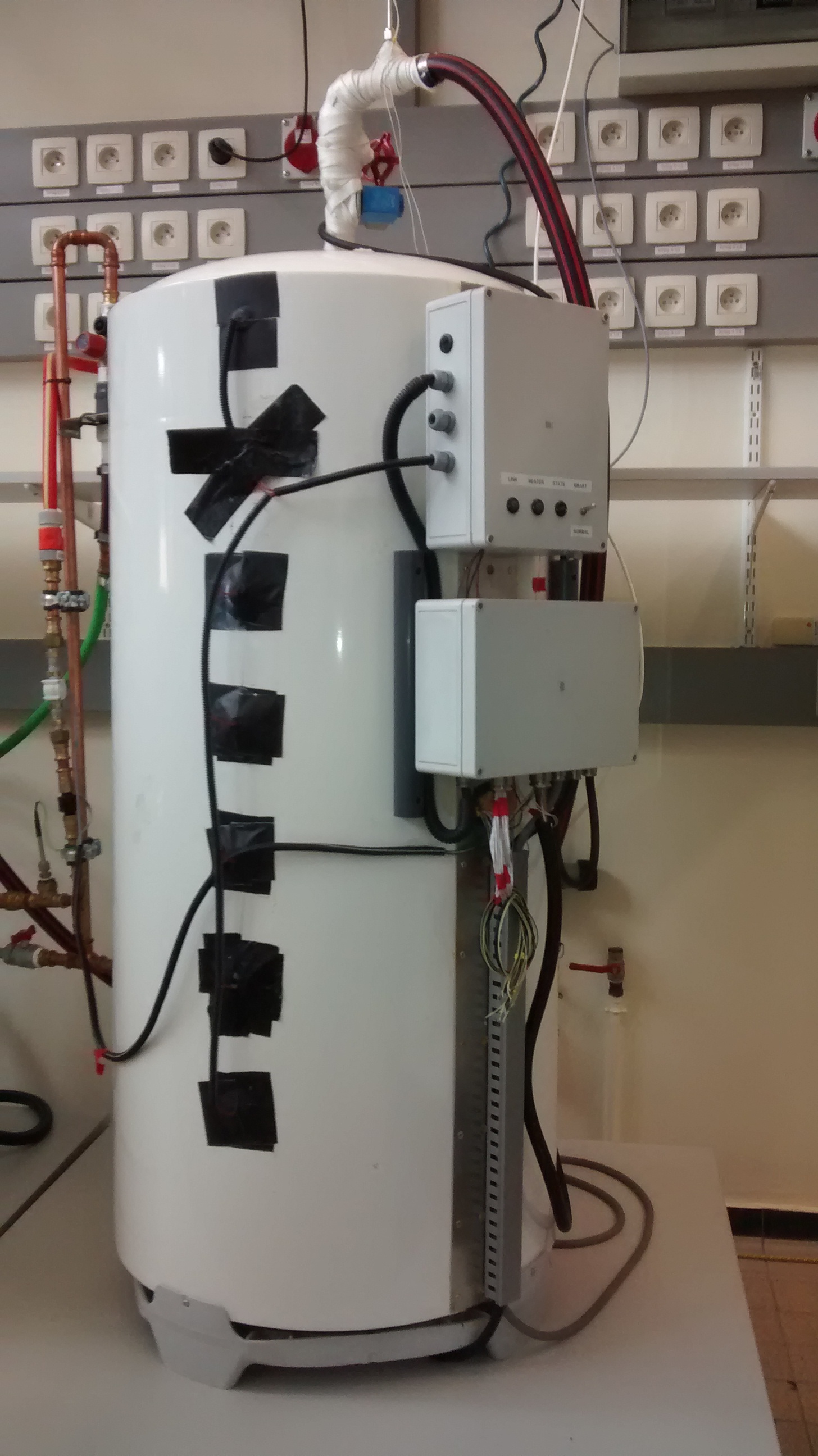}}
\caption{Setup of the electric water heater used during the lab experiment. Eight temperature sensors were placed under the insulation material of the buffer tank. }
\label{boiler_photo}
\end{figure}

This paper proposes a learning agent that minimizes the cost of energy consumption of an electric water heater.
The  agent measures the state of its environment through a set of sensors that are connected along the water buffer of the electric water heater.
However, learning in a high-dimension state space can significantly impact the learning rate of the  RL algorithm.
This is known as the  ``curse of dimensionality".
A popular approach to mitigate its effects is to reduce the dimensionality of the state space during a pre-processing step.
Inspired by the work of~\cite{lange2010deep}, this paper applies an auto-encoder network to reduce the dimensionality of the state vector.
By so doing, this paper makes following contributions:
(1) we demonstrate how a well-established RL technique, fitted Q-iteration, can be combined with an auto-encoder network to minimize the cost of energy consumption of an electric water heater;
(2) in a simulation-based experiment, we assess the performance of different state representations and batch sizes;
(3) we successfully apply an RL agent to an electric water heater in the lab (Fig.~\ref{boiler_photo}). 

The remainder of this paper is organized as follows.
Section~\ref{RW}  gives a  non-exhaustive literature overview of RL related to demand response.
Section~\ref{problem_formulation} maps  the considered demand response problem to  a Markov decision process. 
Section~\ref{feature_selection} describes how an auto-encoder network can be used to find a low-dimensional state representation, followed by a description of the fitted Q-iteration algorithm in Section~\ref{fitted_Q_iteration}.
Section~\ref{results_sim} presents the results of the simulation-based experiments and Section~\ref{results_lab} presents the results of the lab experiment.
Finally, Section~\ref{conclusion} draws  conclusions and discusses future work.

\section{Reinforcement learning}
\label{RW}
This section gives a non-exhaustive overview of recent developments related to Reinforcement Learning (RL) and demand response.
Perhaps the most widely used model-free RL technique applied to a demand response setting is standard Q-learning~\cite{kara2012using,henze2003evaluation,WenZhen,MarinaPscc2014}.
After each interaction with the environment, Q-learning uses  temporal difference learning~\cite{sutton1998reinforcement} to update its state-action value function or Q-function.
A major drawback of Q-learning is that the given observation is discarded  after each update.
As a result, more interactions are needed to spread already known information through the state space.
This inefficient use of information limits the application of Q-learning to real-world applications.

In contrast to  Q-learning, batch RL techniques~\cite{ernst2005tree,riedmiller2005neural,deepmind} are more data efficient, since they store and reuse past interactions.
As a result, batch RL techniques require less interactions with their environment, which makes them  more practical for real-world applications, such as demand response.
%A first batch RL method, experience replay, extends Q-learning and allows it to store and reuse pas observations~\cite{replay}.
Perhaps the most popular batch RL technique which has been applied to a wide range of applications~\cite{lange2010deep,riedmiller2009reinforcement,fonteneau2008variable} is fitted Q-iteration developed by Ernst \etal ~\cite{ernst2005tree}.
Fitted Q-iteration iteratively estimates the Q-function given a fixed batch  of past interactions.
An online version that uses a neural network, neural fitted Q-iteration, has been proposed by Riedmiller \etal in~\cite{riedmiller2005neural}.
Finally, an interesting alternative is to combine experience replay to an incremental RL technique such as Q-learning or SARSA~\cite{adam2012experience}.
%However, when applied to stochastic environments, such as a demand response setting,  Q-learning and fitted Q-iteration can suffer from an overestimation bias of the Q-function caused by the maximization of the expected future reward.
%A promising extension to mitigate this bias is to use a double estimator  as presented in~\cite{DoubleFQI}.
In~\cite{RuelensBRLCluster}, the authors demonstrate how fitted Q-iteration can be used to control a cluster of electric water heaters.
The results indicate that  fitted Q-iteration was able to reduce the cost of energy consumption of a cluster of 100 electric water heaters after a learning period of 40 days.
In addition,~\cite{RuelensBRLDevice} shows how fitted Q-iteration can be extended to reduce the cost of energy consumption of a heat-pump thermostat given that a forecast of the outside temperature is provided.

A promising alternative to the previously mentioned model-free techniques are model-based or model-assisted RL techniques.
For example, the authors of~\cite{PILCO} present a model-based policy search method that learns a Gaussian process to model uncertainties.
In addition, inspired by~\cite{MABRL}, the authors of~\cite{GiuArxive} demonstrate how a model-assisted batch RL technique can be applied to control a building heating system.

\section{Problem formulation}
\label{problem_formulation}
The aim of this work is to develop a controller or agent that minimizes the  cost of energy consumption of an electric water heater, given an external price profile.
This price profile is provided to the agent at the start of each day.
The  agent can measure the temperature of the water buffer through a set of temperature sensors that are connected along the hull of the buffer tank.
Following the approach presented in~\cite{RuelensBRLDevice}, the electric water heater is equipped with a backup controller 
that overrules the control action from the agent when the  safety or comfort constraints of the end user are violated.
A challenge in developing such an agent is that the dynamics of the electric water heater, the future hot water demand and the settings of the backup controller are unknown to the agent.
To overcome this challenge, this paper leverages on the previous work of~\cite{lange2010deep,ernst2005tree,RuelensBRLDevice} and applies techniques from RL.

\subsection{Markov decision process framework}
To apply RL, this paper formulates the underlying sequential decision-making problem of the learning agent as a Markov decision process formulation.
The Markov decision process formulation is defined by its $d$-dimensional state space $X \subset \Rea^{d}$, its action space $U \subset \Rea$, its stochastic discrete-time transition function $f$, and its cost function $\rho$.
The optimization horizon is considered finite, comprising $T \in \Nat\setminus \{0\}$ steps, where at each discrete time step $k$, the state evolves following:
\begin{equation}
\bm{x}_{k+1} = f(\bm{x}_{k},u_{k},\bm{w}_{k}) ~~ \forall k \in \{1,...,T-1\},
\label{eq.f}
\end{equation} where a random disturbance $\bm{w}_{k}$ is drawn from a conditional distribution $p_{\mathcal{W}}(\cdot|\bm{x}_{k})$,
%(for simplicity we will here use $P_{w}(\cdot)$)
 $u_{k} \in U$ is the control action and $\bm{x}_{k} \in X$ the state.
Associated with each state transition, a cost $c_{k}$ is given by: 
\begin{equation}
c_{k}=\rho(\bm{x}_{k},u_{k},\bm{w}_{k}) ~~ \forall k \in \{1,...,T\}.
\label{eq.cost}
\end{equation}
The goal of the learning agent is to find an optimal  control policy ${h^{*}:~X \rightarrow U}$ that minimizes the expected $T$-stage return for any state in the state space.
The expected $T$-stage return  $J^{h}_{T}$ starting from $\bm{x}_{1}$ and following a policy $h$ is defined as follows:
\begin{equation}
J^{h}_{T}(\bm{x}_{1}) =  \underset{p_{\mathcal{W}}(\cdot|\bm{x}_{k})}{\Exp} \left[  \sum_{k=1}^{T}{\rho(\bm{x}_{k},h(\bm{x}_{k}),\bm{w}_{k})}\right],
\end{equation}
where $\Exp$ denotes the expectation operator over all possible stochastic realizations.

A more convenient way to  characterize a policy is  by using  a state-action value function or Q-function:
\begin{equation}
Q^{h}(\bm{x},u) = \underset{p_{\mathcal{W}}(\cdot|\bm{x})}{\Exp} \left[\rho(\bm{x},u,\bm{w}) + J^{h}_{T}(f(\bm{x},h(\bm{x}),\bm{w})) \right],
\label{Qfunction}
\end{equation}
which indicates the cumulative return starting from  state $\bm{x}$ and by taking action $u$ and following $h$ thereafter.

The optimal Q-function corresponds the best Q-function that can be obtained by any policy:
\begin{equation}
Q^{*}(\bm{x},u) = \underset{h}{\text{min~}} Q^{h}(\bm{x},u).
\end{equation}
Starting from an optimal Q-function for every state-action pair, the optimal policy $h^{*}$ is calculated as follows:
\begin{equation}
h^{*}(\bm{x})  \in \underset{u \in U}{\text{arg min~}} Q^{*}(\bm{x},u),
\label{Qpolicy}
\end{equation}
where $Q^{*}$ satisfies the Bellman optimality equation~\cite{BellmanDP}:
\begin{align}
Q^{*}(\bm{x},u) = \underset{w\sim\probw}{\Exp}\left[\rho(\bm{x},u,\bm{w}) +  \underset{u' \in U}{\text{min~}} Q^{*}(f(\bm{x},u,\bm{w}),u') \right].
\label{Qfunction}
\end{align}

Following the notation introduced in~\cite{RuelensBRLDevice}, the next three paragraphs give a description of the state,  the action, and the  cost function tailored to an electric water heater.

\subsection{Observable state vector}
\label{sec:StateDescription}
The observable state vector of an electric water heater contains a time-related component and a controllable  component.
The time-related component $\bm{x}^{\textrm{t}}$ describes the part of the state related to timing, which is relevant for the dynamics of the system.
Specifically, the tap water demand of the end user is considered to have diurnal and weekly patterns.
As such, the time-related component contains the day of the week and the quarter in the day.
The controllable component $\bm{x}^{\textrm{ph}}$ represents physical state information that is measured locally and is influenced by the control action. 
The controllable component contains the temperature measurements of the $n_{\mathrm{s}}$ sensors  that are connected along the hull of the storage tank.
The observable state vector is given by:
\begin{equation}
\bm{x}_{k} = (\underbrace{d,t}_{\bm{x}^{\textrm{t}}_{k}},\underbrace{T^{1}_{k},\ldots,T^{i}_{k},\ldots,T^{n_{\mathrm{s}}}_{k}}_{\bm{x}^{\textrm{ph}}_{k}}),\\
\label{original_state}
\end{equation}
where  $d\in\{1,\ldots,7\}$ is the current day of the week, $t\in\{1,\ldots,96\}$ is the quarter in the day and $T^{i}_{k}$ denotes the temperature measurements of sensor $i$ at time step $k$.

\subsection{Control action}
The learning agent can control the heating element of the electric water heater with a binary control action $u_{k} \in \{0,1\}$, where 0 indicates off and 1 on.
However, the backup mechanism, which enacts the comfort and safety constraints of the end user, can overrule this  control action of the learning agent.
The function  $B:X \times U \rightarrow U^{\mathrm{ph}}$ maps the control action $u_{k} \in U$ to a physical  action $\uphk \in U^{\mathrm{ph}}$ according to:
\begin{equation}
\uphk = B(x_{k},u_{k},\bm{\theta}),
\end{equation} 
where the vector $\bm{\theta}$ defines  the safety and user-defined comfort settings of the backup controller.
In order to have a generic approach we assume that the logic of the backup controller is unknown to the learning agent.
However, the learning agent can measure the  physical  action $\uphk$ enforced by the backup controller (see Fig.~\ref{boiler_auto_encoder}), which is required to calculate the cost.

The logic of the backup controller of the electric  water heater is defined as: 
\begin{align}
B(\bm{x},u,\bm{\theta}) = \left\{\begin{matrix} P^{\mathrm{e}}&\text{if}&\soc(\bm{x},\bm{\theta}) \leq \socmin(\bm{\theta})  \\  
uP^{\mathrm{e}}&\text{if}&\socmin(\bm{\theta}) <\soc(\bm{x},\bm{\theta})<\socmax(\bm{\theta}) ,\\  
0&\text{if}&\soc(\bm{x},\bm{\theta}) \geq \socmax(\bm{\theta})   \end{matrix}\right.
\label{backup_controller}
\end{align} 
where  $P^{\mathrm{e}}$ is the electrical power rating of the heating element, $\soc(\bm{x},\bm{\theta})$ is the current state of charge and  $\socmin(\bm{\theta})$ and $\socmax(\bm{\theta})$ are the upper and lower bounds for the state of charge.
A detailed description of how the state of charge is calculated can by found in~\cite{SMARTBOILER}.

\subsection{Cost function}
At the start of each optimization period $T\Delta t$, the learning agent receives a price vector $\bm{\lambda} = \{\lambda_{k}\}_{k=1}^T$ for the next $T$ time steps. % for the optimization period $T \Delta t$, with $\Delta t $  the length of one control period. 
At each time step, the agent receives a cost $c_{k}$ according to:
\begin{equation}
c_{k}=  \uphk\lambda_{k}\Delta t,
\label{eq_ToU}
\end{equation}
where $\lambda_{k}$ is the electricity price during time step $k$, and $\Delta t $  the length of one control period.

\section{Batch of four-tuples}
\label{feature_selection}
Generally, batch RL techniques estimate the Q-function based on a batch of four-tuples $(\bm{x}_{l},u_{l},\bm{x}_{l}',c_{l})$.
This paper, however, considers the following batch of four-tuples:
\begin{equation}
\mathcal{F} = \{(\bm{x}_{l},u_{l},\bm{x}_{l}',\uphl), l = 1,\ldots, \#\mathcal{F}\},
\end{equation}
where for each $l$, the next state $\bm{x}_{l}'$, and the physical action $\uphl$ have been obtained as a result of taking control action $u_{l}$ in the state $\bm{x}_{l}$.
Note that, $\mathcal{F}$ does not include the observed cost $c_{l}$, since the cost depends on the price vector that is provided to the learning agent at the start of each day.

As defined by~(\ref{original_state}), $\bm{x}_{l}$ contains all temperature measurements of the sensors connected to the hull of the water buffer.
Learning in a high-dimensional state space  requires more observations from the environment to estimate the Q-function, as more tuples are needed to cover the state-action space. 
This  is known as the ``curse of dimensionality".
This curse becomes even more pronounced  in practical applications, where each observation corresponds to a ``real" interation with the environment.

A pre-processing step can be used to find a compact and more efficient representation of the state space and can help to converge to a good policy much faster~\cite{tim_brys}.
A popular technique to find a compact representation is to use a handcrafted feature vector based on insights of the considered control problem~\cite{bertsekas1996neuro}. 
Alternative approaches that do not require prior knowledge are unsupervised feature learning algorithms, such as auto-encoders~\cite{lange2010deep} or a principal component analysis~\cite{tim_brys}.

\begin{figure}[t]
\centerline{\includegraphics[width=1\columnwidth]{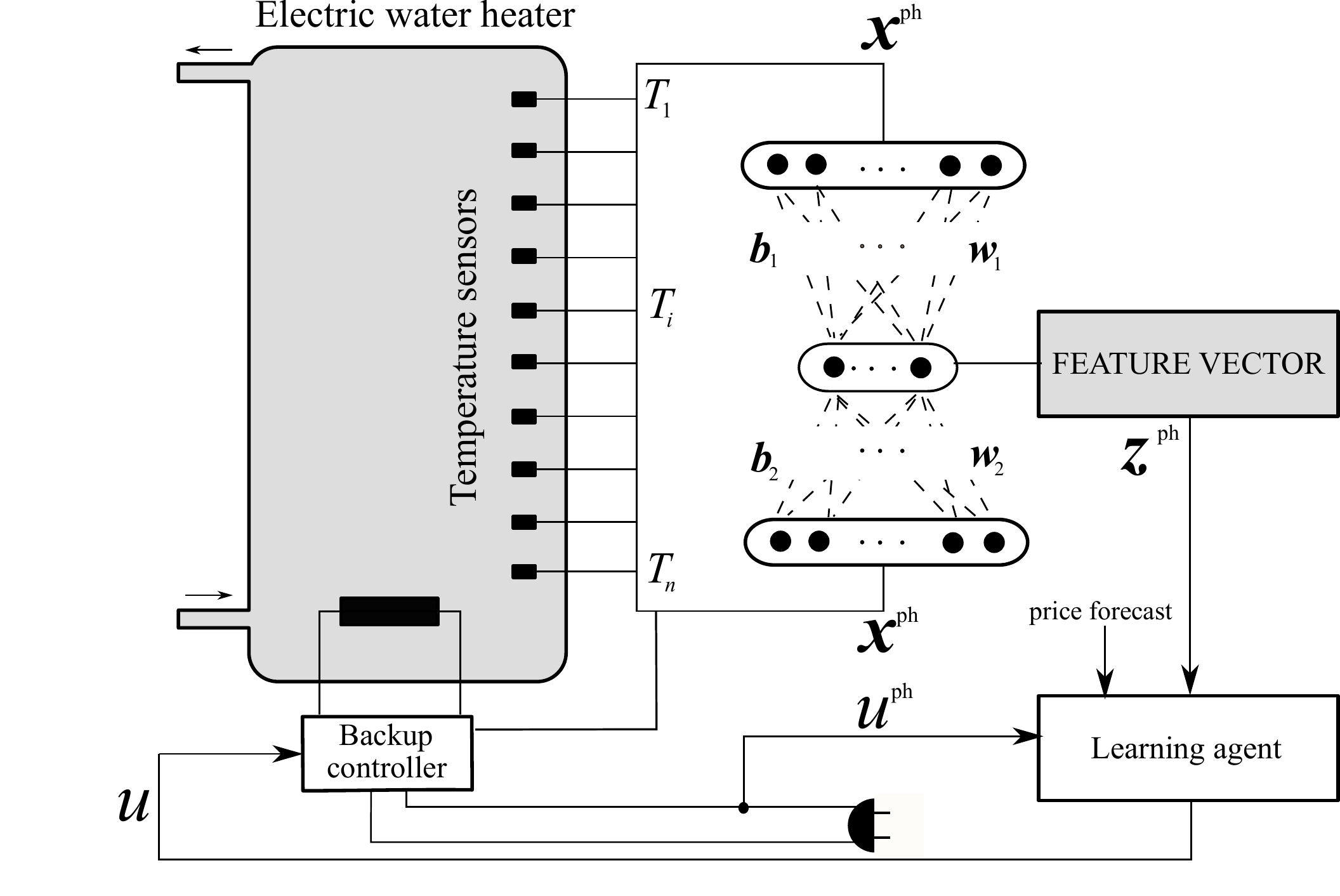}}
\caption{Setup of the simulation-based experiment. An auto-encoder network is used to find a compact representation of the temperature measurements.}
\label{boiler_auto_encoder}
\end{figure}

As illustrated in Fig.~\ref{boiler_auto_encoder}, this paper demonstrates how an auto-encoder can be used to find a compact representation of the sensory input data.
An auto-encoder network is a neural network that maps its output back to its input.
By selecting a lower number of neurons in the middle hidden layer than in the input layer $p<d$, the auto-encoder can be used to reduce the dimensionality of the input data.

The reduced feature vector $\bm{z}_{l} \in Z \subset \Rea^{p}$ can be calculated as follows:
\begin{equation}
\label{encodedstate}
\bm{z}_{l} = (\bm{x}^{\textrm{t}}_{l},{\Phi}_{\mathrm{enc}}(\bm{x}^{\textrm{ph}}_{l},\bm{w},\bm{b})),
\end{equation}
%with
%\begin{equation}
%\label{encodedstate}
%\bm{z}_{l}^{\textrm{ph}} = {\Phi}_{\mathrm{enc}}(\bm{x}^{\textrm{ph}}_{l},\bm{w},\bm{b}),
%\end{equation}
where $\bm{w}$ and $\bm{b}$ denote the weights and the biases that connect the input layer with the middle hidden layer of the auto-encoder network.
The function ${\Phi}_{\mathrm{enc} }:X \rightarrow Z$ is an encoder function and maps the observed state vector $\bm{x}_{l}$ to the feature vector $\bm{z}_{l}$.
%A detailed description of the training algorithm is out of the scope of this paper. 
To train the weights of the auto-encoder, a conjugate gradient descent algorithm is used as presented in~\cite{Scholz}.

In the next section, fitted Q-iteration is used to find a policy ${h :Z \rightarrow U}$ that maps every feature vector to a control action using batch $\mathcal{R}$:
\begin{equation}
\mathcal{R} = \{(\bm{z}_{l},u_{l},\bm{z}_{l}',\uphl), l = 1,\ldots, \#\mathcal{R}\},
\end{equation}
which consists of feature vectors with a dimensionality $p$.

Since we apply the auto-encoder on the input data of the supervised learning algorithm, we assume that all input data is equally important.
As such, it is possible that we ignore low-variance yet potentially useful components during the learning process.
A possible route of future work would be to add a regularization term to the regression algorithm of the supervised learning algorithm to prevent the risk of overfitting without the risk of ignoring potentially important data.
%\subsection{Feature selection}
%\label{sec:PolicyRepair}

\begin{algorithm}[t]
\caption{Fitted Q-iteration~\cite{ernst2005tree} using feature vectors.}
\label{FQI}
\begin{algorithmic}[1] 
\algsetup{linenosize=\tiny}
\REQUIRE $\mathcal{R} =\left\{(\bm{z}_{l},u_{l},\bm{z}_{l}',\uphl),l=1,\ldots,\#\mathcal{R} \right\}, \{\lambda_{t}\}_{t=1}^T$
\STATE initialize $\widehat{Q}_{0}$ to zero
\FOR {$N = 1,\ldots,T$}
\FOR {$l = 1,\ldots,\#\mathcal{R}$}
%\STATE $c_{l} \leftarrow \uphl \lambda_{t} \vartriangleright$ \textit{where $t$ corresponds to the  time-related component $\bm{x}^{\textrm{t}}_{l}=(d,t)$  of state $\bm{z}_{l}$} 
\STATE $c_{l} \leftarrow \uphl \lambda_{t} \vartriangleright$ \textit{where $t$ is to the quarter in the day of the time-related component $\bm{x}^{\textrm{t}}_{l}=(d,t)$  of state $\bm{z}_{l}$} 
\label{cost_fqi}
\STATE $Q_{N,l}\leftarrow c_{l}+\underset{u \in U}{\text{min~}} \widehat{Q}_{N-1}(\bm{z}_{l}',u) $
\label{Q_function}
\ENDFOR
\STATE use a regression technique to obtain $\widehat{Q}_{N}$ from \text{~~~~~~~~~} $\mathcal{T}_{\mathrm{reg}} =\left\{\big(\left(\bm{z}_{l},u_{l}\right),Q_{N,l}\big),l =1,\ldots,\#\mathcal{R}\right\}$.
\ENDFOR
\ENSURE $\widehat{Q}^{*}=\widehat{Q}_{T}$
\end{algorithmic}
\end{algorithm}

\section{Fitted Q-iteration}
\label{fitted_Q_iteration}
This section describes the learning algorithm and  the exploration strategy of the agent based on the batch of feature vectors $\mathcal{R}$ presented in the previous section.

\subsection{Fitted Q-iteration }
Fitted Q-iteration iteratively builds a training set $\mathcal{T}_{\mathrm{reg}}$ with all  state-action pairs $(\bm{z},u)$ in $\mathcal{R}$  as the input.
The target values consist of the corresponding Q-values, based on the approximation of the  Q-function of the previous iteration.
In the first iteration ($N=1$), the Q-values  approximate the expected cost (line~\ref{Q_function} in Algorithm 1). 
In the subsequent iterations, Q-values are updated using the Q-function of the previous iteration. 
As a result, Algorithm 1 needs $T$ iterations until the Q-function contains all information about the future costs. 
Note that,  the cost corresponding to each tuple  is recalculated using  price vector $\bm{\lambda}$ that is provided at the start of the day (line~\ref{cost_fqi} in Algorithm~\ref{FQI}).
As a result, the algorithm can reuse past experiences to find a control policy for the next day.
Following~\cite{ernst2005tree}, Algorithm~\ref{FQI} applies an ensemble of extremely randomized trees as a regression algorithm to estimate the Q-function.
An empirical study of the accuracy and convergence properties of extremely randomized trees can be found in~\cite{ernst2005tree}.
However, in principle, any regression algorithm, such as neural networks~\cite{deepmind,riedmiller2009reinforcement}, can be used to estimate the Q-function.

\subsection{Boltzmann exploration}
During the day, the learning agent uses a Boltzmann exploration strategy~\cite{kaelbling1996reinforcement} and  selects an action with the following probability:
\begin{equation}
P\left(u|\bm{z} \right) = \frac{e^{\widehat{Q}^{*}\left(\bm{z},u\right)/\tau_{d}}}{\Sigma_{u'\in U} e^{\widehat{Q}^{*}\left(\bm{z},u'\right)/\tau_{d}}},
\label{Boltzmann_ex}
\end{equation}
where $\tau_{d}$ is the Boltzmann temperature at day $d$,  $\widehat{Q}^{*}$ is the Q-function from Algorithm~\ref{FQI} and $\bm{z}$ is the current feature vector measured by the learning agent.
If $\tau_{d} \rightarrow 0$, the exploration  will decrease and the policy will become greedier. 
Thus by starting with a high $\tau_{d}$ the exploration starts completely random, however as $\tau_{d}$ decreases the policy directs itself to the most interesting state-action pairs.
In the evaluation experiments, $\widehat{Q}^{*}$ in~(\ref{Boltzmann_ex}) is linearly scaled between $[0,100]$ and the $\tau_{1}$ is set to $100$ at the start of the experiment, which will result in an equal probability for all actions.
The Boltzmann temperature is updated as follows $\tau_{d} = \tau_{d-1} - \Delta \tau$, which increases the probability of selecting higher valued actions.

%Since this  Q-function is based on a batch of feature dvectors,  the current measured state vector needs to be projected to its low-dimensional representation (\ref{encodedstate}).
%The parameter $\tau_{k}$ is decreased during the experiment and the policy becomes greedier as more tuples are added to the batch.
%Note that any exploration strategy, such as $\varepsilon$-greedy can be used in combination with fitted Q-iteration.
%An interesting property of the Boltzmann exploration is that it uses information captured in the Q-function to explore interesting states.

\section{Simulation-based Results}
\label{results_sim}
This section describes the results of the simulation-based experiments, which use a non-linear stratified tank model with \layers temperature layers.
A detailed description of the stratified tank model can be found in~\cite{SMARTBOILER}.
The specifications of the electric water heater are chosen in correspondence with the electric water heater used during the lab experiment (see Section~\ref{results_lab}).
The simulated electric water heater has a power rating of 2.36kW and has a water buffer of 200 liter. 
The experiments use realistic hot water profiles with a mean daily consumption of \liters liter~\cite{jordan2001realistic} and use price information from the Belgian day-ahead~\cite{belpex} and balancing market~\cite{elia}. 
The learning agent can measure the temperature of the \layers temperature layers obtained with the simulation model.

The aim of the first simulation-based experiment is to find a compact  state representation using an auto-encoder network and to assess the impact of the state representation on the performance of fitted Q-iteration.
The second simulation-based experiment compares the result of fitted Q-iteration with the default thermostat controller.

\begin{figure}[t]
\centerline{\includegraphics[width=\columnwidth]{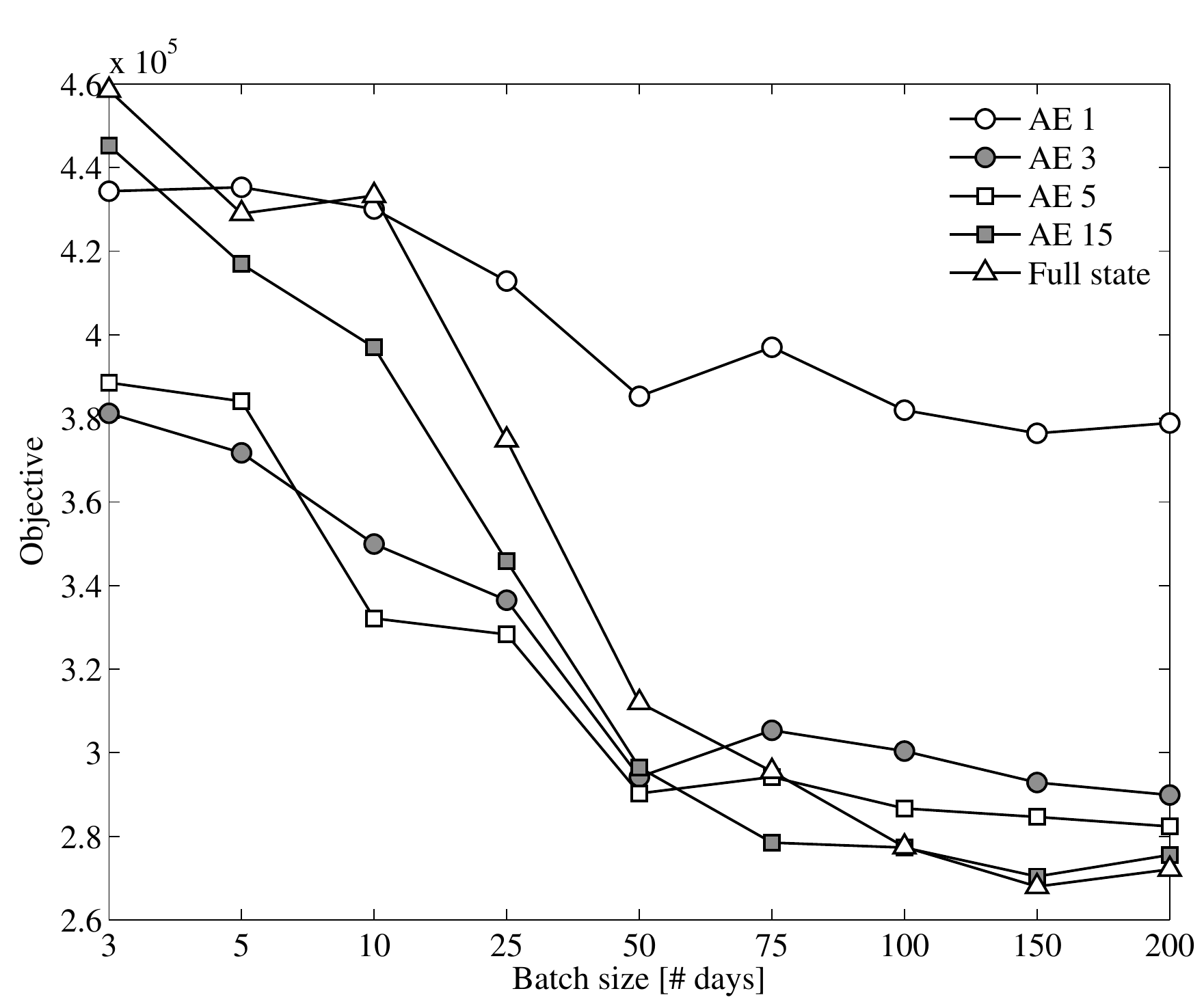}}
\caption{Simulation-based results of fitted Q-iteration using five state representations and different batch sizes. 
The full state contains 50 temperature measurements.
A non-linear dimensionality reduction  with Auto-Encoder (AE) is used to find a compact representation of the temperature measurements. 
Each  marker point represents the average result of \simrun simulation runs.}
\label{AE_plot}
\end{figure}

\subsection{Step 1: feature selection}
This experiment compares the performance of  fitted Q-iteration combined with different  feature representations for different fixed batch sizes.
An auto-encoder (AE) network that reduces the original sensory input vector (50 temperature sensors) to 5 dimensions is denoted by AE 5.
The simulations are repeated for \simrun  simulation days.
The average cost of energy consumption of these \simrun simulations is depicted in Fig.~\ref{AE_plot}.
As can be seen in Fig.~\ref{AE_plot}, the performance of fitted Q-iteration combined with a specific state representation depends on the number of tuples in the batch.
For example for a batch size of 10 days,  AE 3 results in a lower cost than AE 15, while after 75 days, AE 15 will result in a lower cost than AE 3.
In addition, as can be seen from Fig.~\ref{AE_plot}, AE 1 resulted in a relatively bad policy, independent of the batch size.

In general, it  can be concluded that for a batch of limited size, fitted Q-iteration with a low-dimensional feature vector will outperform  fitted Q-iteration using the full state information, i.e. 50 temperature measurements.
By learning in a low-dimensional state space, it is possible to learn with a smaller and more efficient representation.
As a result, the agent requires less observations to converge to a better control policy than when the full state information is used.
In addition, as more observations will result in a more efficient coverage of the state-action space,  it can be seen from Fig.~\ref{AE_plot} that the result of fitted Q-iteration with the full state improves significantly as the batch size increases.
In the following subsection, we present the results of AE 5 in more detail.
A  method for selecting an appropriate feature representation during the learning process will be part of our future work (Section~\ref{conclusion}).

%These results indicate that a low
%The online selection of the best state reduction is will be part of our future research.
%In the Section~\ref{conclusion}, we will discuss a possible approach to select an optimal feature reduction.

%\begin{figure}[t]
%\centerline{\includegraphics[width=\columnwidth]{cumulative_cost_sim}}
%\caption{Simulation results using fitted Q-iteration combined with an auto-encoder and the default thermostat controller.}
%\label{boiler_sim_performance}
%\end{figure}

\begin{figure}[t]
\centering
\begin{picture}(100,215)
\put(-78,0){\includegraphics[width=9cm]{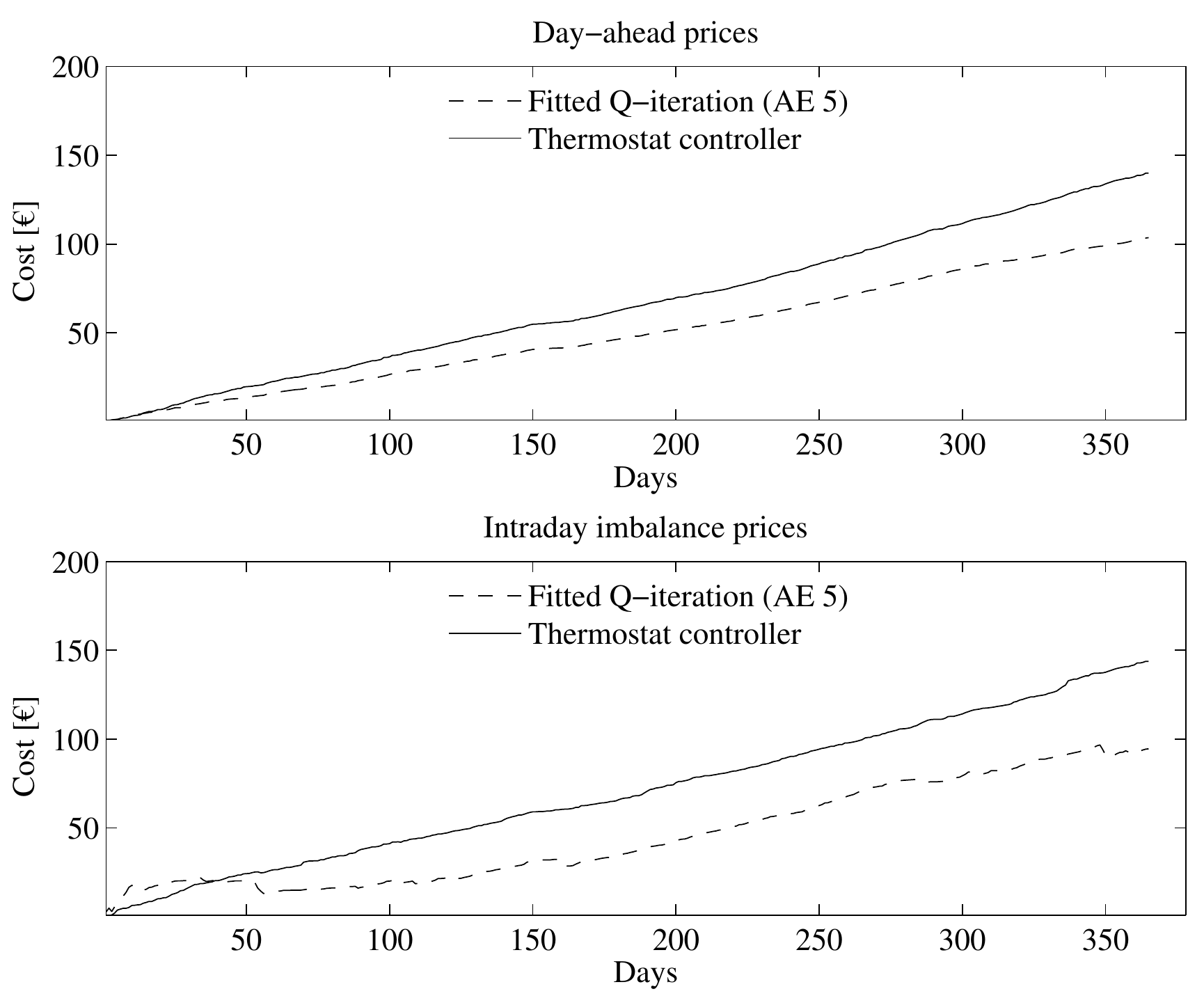}}
\put(-75,195){\textbf{a}}
\put(-75,90){\textbf{b}}
\end{picture}
\caption{Cumulative energy cost of fitted Q-iteration with a non-linear dimensionality reduction and of the thermostat controller.
 Results for one year using day-ahead prices (\textbf{a})  and imbalance prices (\textbf{b}).}
\label{boiler_sim_performance}
\end{figure}

\subsection{Step 2: evaluation}
Fig.~\ref{boiler_sim_performance} compares the total cost of energy consumption using fitted Q-iteration combined with AE 5 against the default thermostat controller for two relevant price profiles, i.e. day-ahead prices (top plot) and  imbalance prices  (bottom plot).
The default thermostat controller enables the heating element when the state-of-charge drops below its minimum threshold and stays enabled until the state-of-charge reaches 100$\%$. Note that in contrast to the learning agent, the default controller is agnostic about the price profile.
 
The experiment starts with an empty batch and the tuples of the current day are added to the given batch at the end of each day.
At the start of each day, the auto-encoder is trained to find a batch of  compact feature vectors, which are then  used by fitted Q-iteration to estimate the Q-function for the next day.
Online, the learning agent uses a Boltzmann exploration strategy with $\Delta \tau$ set  to 10, which results in 10 days of exploration.

The results of the experiment indicate that fitted Q-iteration was able to reduce the total cost of energy consumption  by  \dayahead$\%$ for the day-head prices and  by \balancing$\%$ for the imbalance prices compared to the default strategy.
Note that the  imbalance prices are generally more volatile than the day-ahead prices, as they reflect real-time imbalances due to forecasting errors of renewable energy sources, such as wind and solar, which were not foreseen in the day-ahead market.

%\begin{figure}[t]
%\centerline{\includegraphics[width=9cm]{powers_imb}}
%\caption{Simulation results using fitted Q-iteration combined with an auto-encoder (AE 5) using intraday balancing prices.  
	%The top plot shows the temperature of the layers. The middle plot depicts the price profile and the power profile and the bottom plot depicts the hot tap water demand.}
%\label{snap_shot_imbalance}
%\end{figure}

\begin{figure}[t]
\centering
\begin{picture}(100,200)
\put(-78,0){\includegraphics[width=9cm]{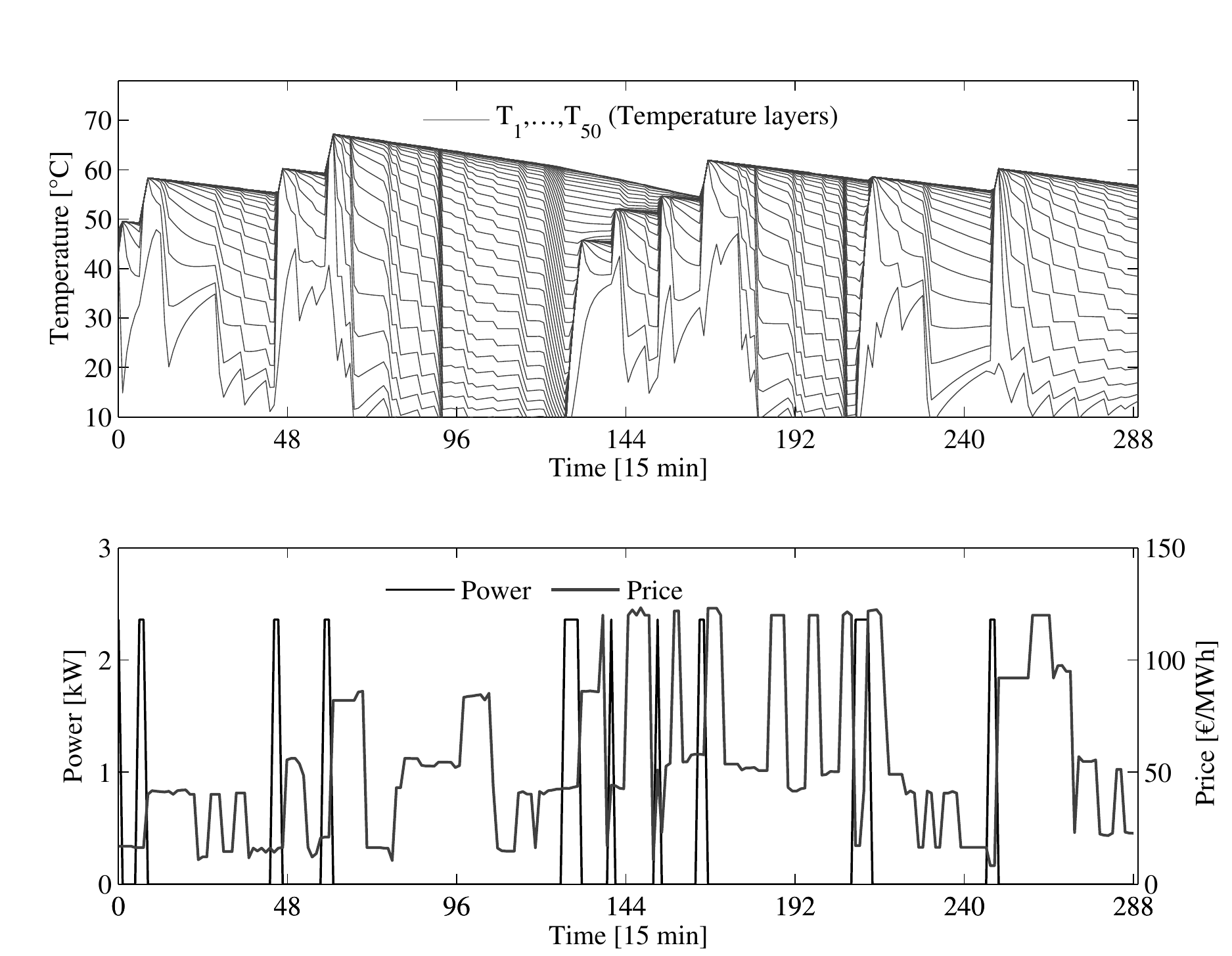}}
\put(-68,180){\textbf{a}}
\put(-68,80){\textbf{b}}
\end{picture}
\caption{Simulation-based results of a mature agent using fitted Q-iteration with a non-linear dimensionality reduction. \textbf{a}, Temperature profiles of the 50 simulation layers. \textbf{b}, Power consumption (black) and  imbalance prices (gray). }
\label{snap_shot_imbalance}
\end{figure}

\begin{figure}[t]
\centering
\begin{picture}(100,200)
\put(-78,0){\includegraphics[width=9cm]{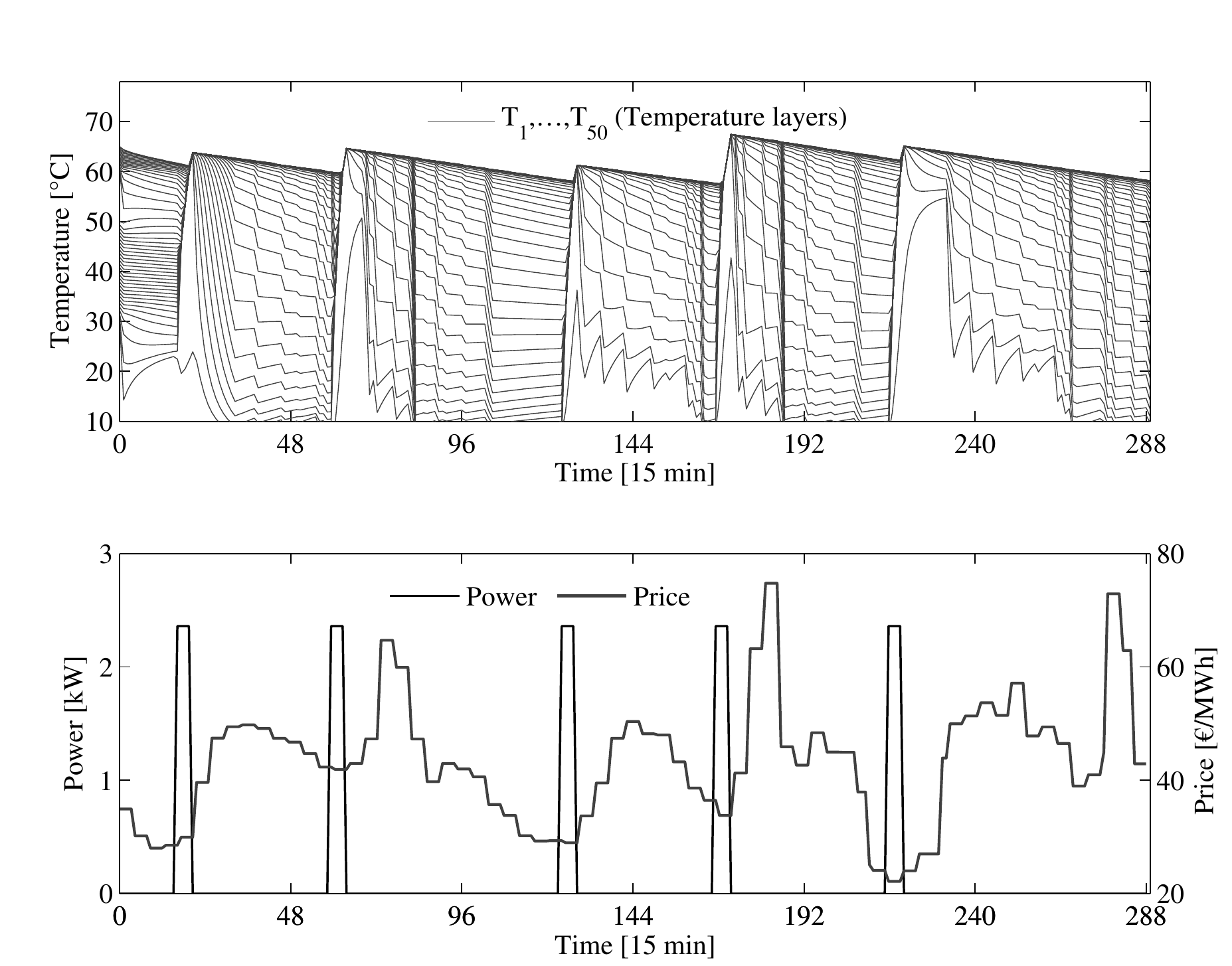}}
\put(-68,180){\textbf{a}}
\put(-68,80){\textbf{b}}
\end{picture}
\caption{Simulation-based results of a mature agent using fitted Q-iteration with a non-linear dimensionality reduction. \textbf{a}, Temperature profiles of the 50 simulation layers. \textbf{b}, Power consumption (black) and day-ahead prices (gray). }
\label{snap_shot_belpex}
\end{figure}

The temperature profiles of the simulation layers and  power profiles of a ``mature" learning agent (batch size of 100 days) for the day-ahead  and imbalance prices are depicted in Fig.~\ref{snap_shot_imbalance}  and Fig.~\ref{snap_shot_belpex}.
It can be seen in the bottom plot of both figures that the mature learning agent reduces the cost of energy consumption by consuming energy during low price moments.

\section{Lab Results}
\label{results_lab}
The aim of our lab experiment is to demonstrate that fitted Q-iteration can be successfully applied to minimize the  cost of energy consumption of a real-world electric water heater.

\subsection{Lab setup}
The setup used in the lab experiment was part of a pilot project on residential demand response in Belgium~\cite{Dupont}, where a cluster of 10 electric water heaters was used for direct load control.  
Fig.~\ref{boiler_photo} shows the electric water heater used during the lab experiment.
The electric water heater is a standard unit that was equipped with eight  temperature sensors and a controllable power relay.
 A controllable valve connected to the outlet of the buffer tank is used to simulate the hot water demand of a household with a mean daily flow volume of \liters liter~\cite{jordan2001realistic}.
An Arduino prototyping platform with a  JSON/RPC 2.0 interface is used to communicate with a computer in the lab\footnote{Intel Core i5 - 4GB Memory}, which runs the learning agent that uses fitted Q-iteration.
Fitted Q-iteration is implemented in Python and Scikit-learn~\cite{Scikit} is used to estimate the Q-function, using an ensemble of extremely randomized trees~\cite{ernst2005tree}.

Similar as in the previous simulation-based experiments, it is assumed that the learning agent is provided with  a deterministic external price profile for the following day.
The learning agent uses a Boltzmann exploration strategy with $\Delta \tau$ set  to 10, which results in 10 days of exploration.
In order to compare the performance of the lab experiment with the performance of the simulation-based experiments, we used both day-ahead prices and imbalance prices.

%\begin{figure}[t]
%\centerline{\includegraphics[width=9cm]{powers_belp}}
%\caption{Simulation results using fitted Q-iteration combined with an auto-encoder (AE 5) using intraday balancing prices.  
	%The top plot shows the temperature of the layers. The middle plot depicts the price profile and the power profile and the bottom plot depicts the hot tap water demand.}
%\label{snap_shot_belpex}
%\end{figure}

\subsection{Evaluation}
The performance of the learning agent was evaluated using different feature vectors as presented in Section~\ref{results_sim}. 
The best performance, however, was obtained by including the eight temperature measurements in the observable state vector. 

\begin{figure}[t]
\centering
\begin{picture}(100,200)
\put(-78,0){\includegraphics[width=9cm]{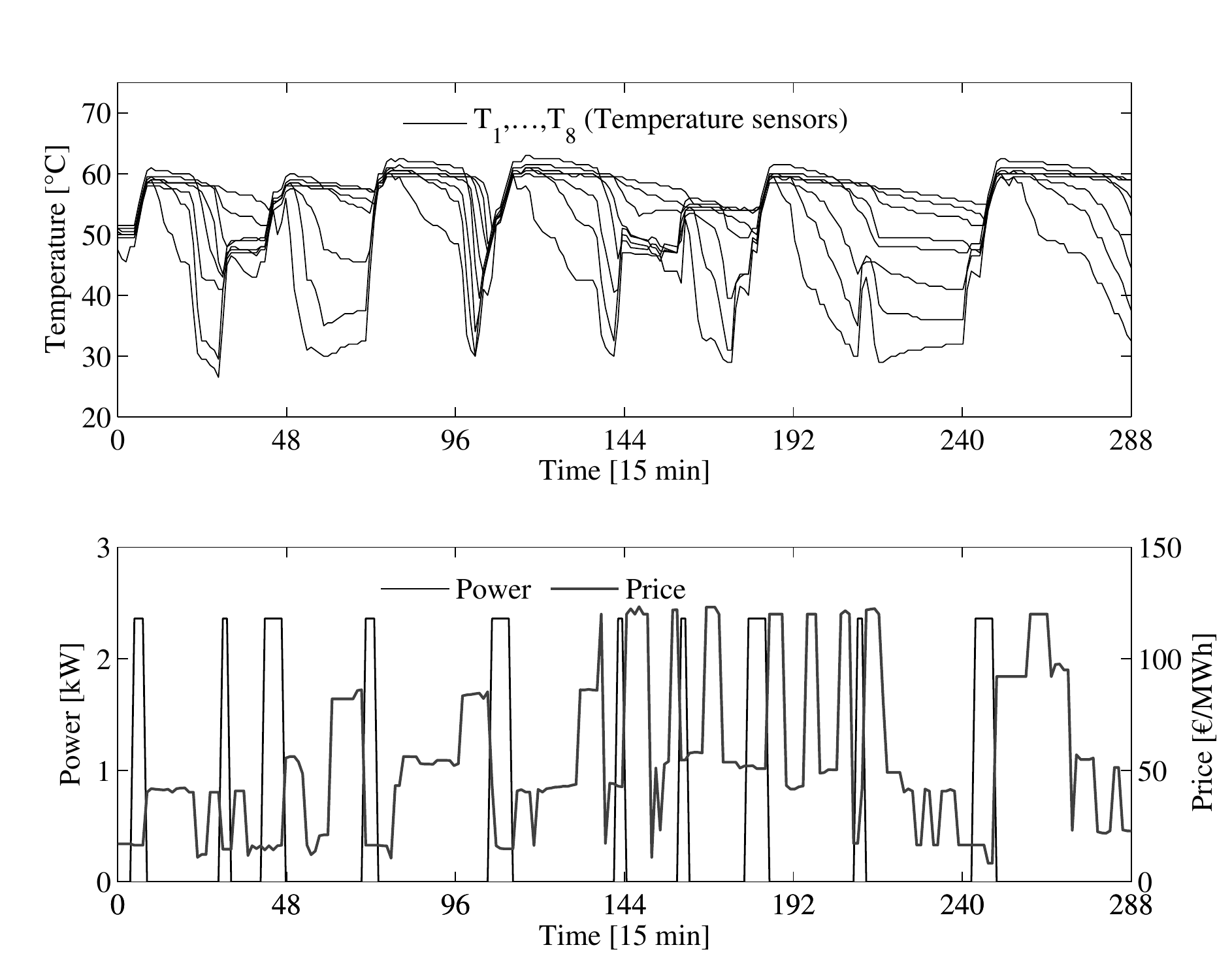}}
\put(-68,180){\textbf{a}}
\put(-68,80){\textbf{b}}
\end{picture}
\caption{Lab results of a mature agent using fitted Q-iteration with the full state (eight temperature measurements). \textbf{a}, Measurements of the temperature sensors. \textbf{b}, Power consumption (black) and  imbalance prices (gray). }
\label{powers_lab_imb}
\end{figure}

\begin{figure}[t]
\centering
\begin{picture}(100,200)
\put(-78,0){\includegraphics[width=9cm]{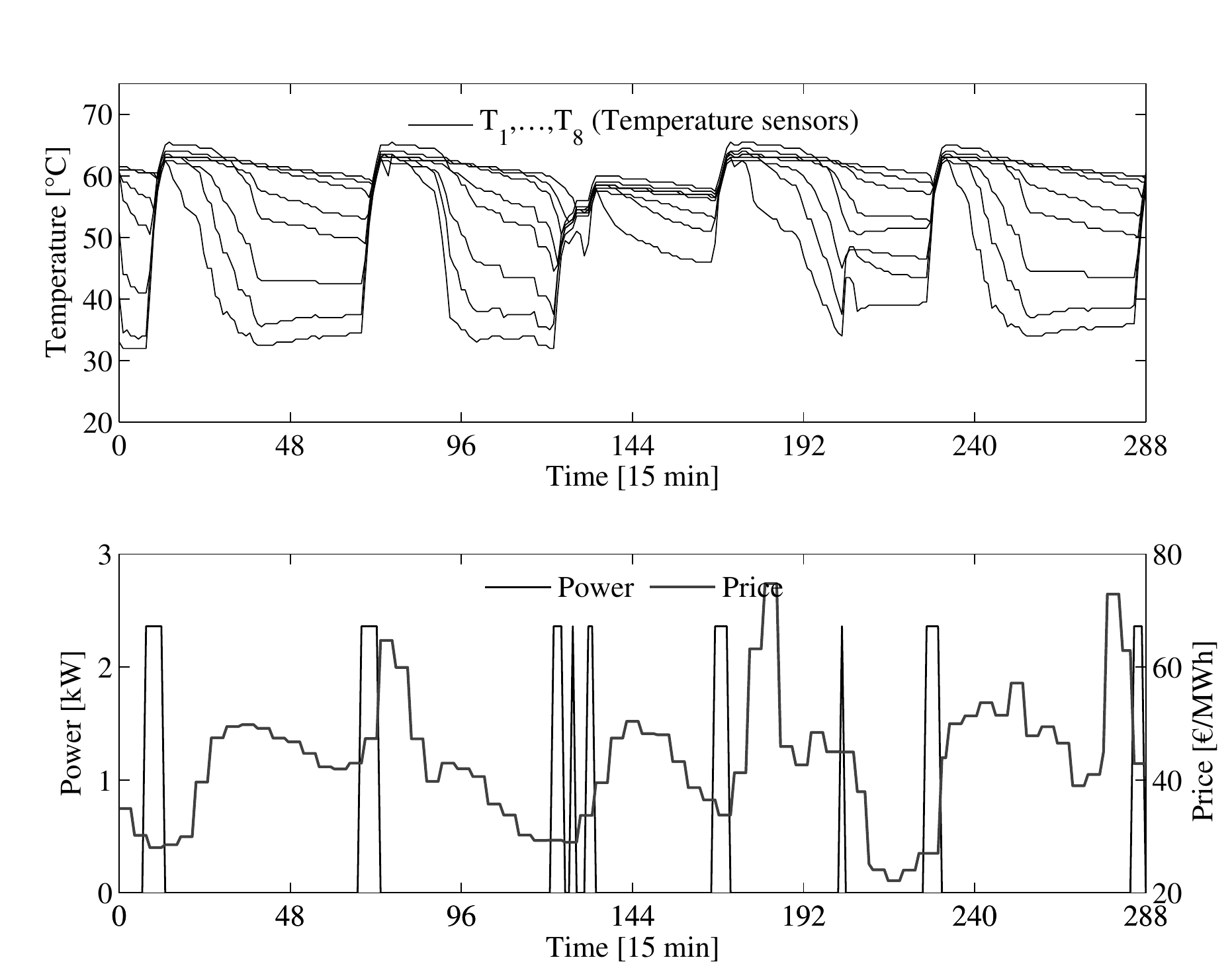}}
\put(-68,180){\textbf{a}}
\put(-68,80){\textbf{b}}
\end{picture}
\caption{Lab results of a mature agent using fitted Q-iteration with the full state (eight temperature measurements). \textbf{a}, Measurements of the temperature sensors. \textbf{b}, Power consumption (black) and day-ahead prices (gray). }
\label{powers_lab_belp}
\end{figure}

Using identical price profiles and tap water demands, Fig.~\ref{powers_lab_imb} and Fig.~\ref{powers_lab_belp} show the temperature measurements and the power profiles of the mature learning agent using imbalance prices and day-ahead prices.
As can be seen, the learning agent was able to successfully minimize the cost of energy consumption by consuming during low price moments.
%However, due to the stochastic behavior of the end user, i.e. tap water demand, it can occur that the electric water heater consumes energy during high prices moments.
%This consumption is  triggered by the backup controller, which guarantees the comfort of the end user, independent of action of the learning agent.

\begin{figure}[t]
\centering
\begin{picture}(100,200)
\put(-78,0){\includegraphics[width=9cm]{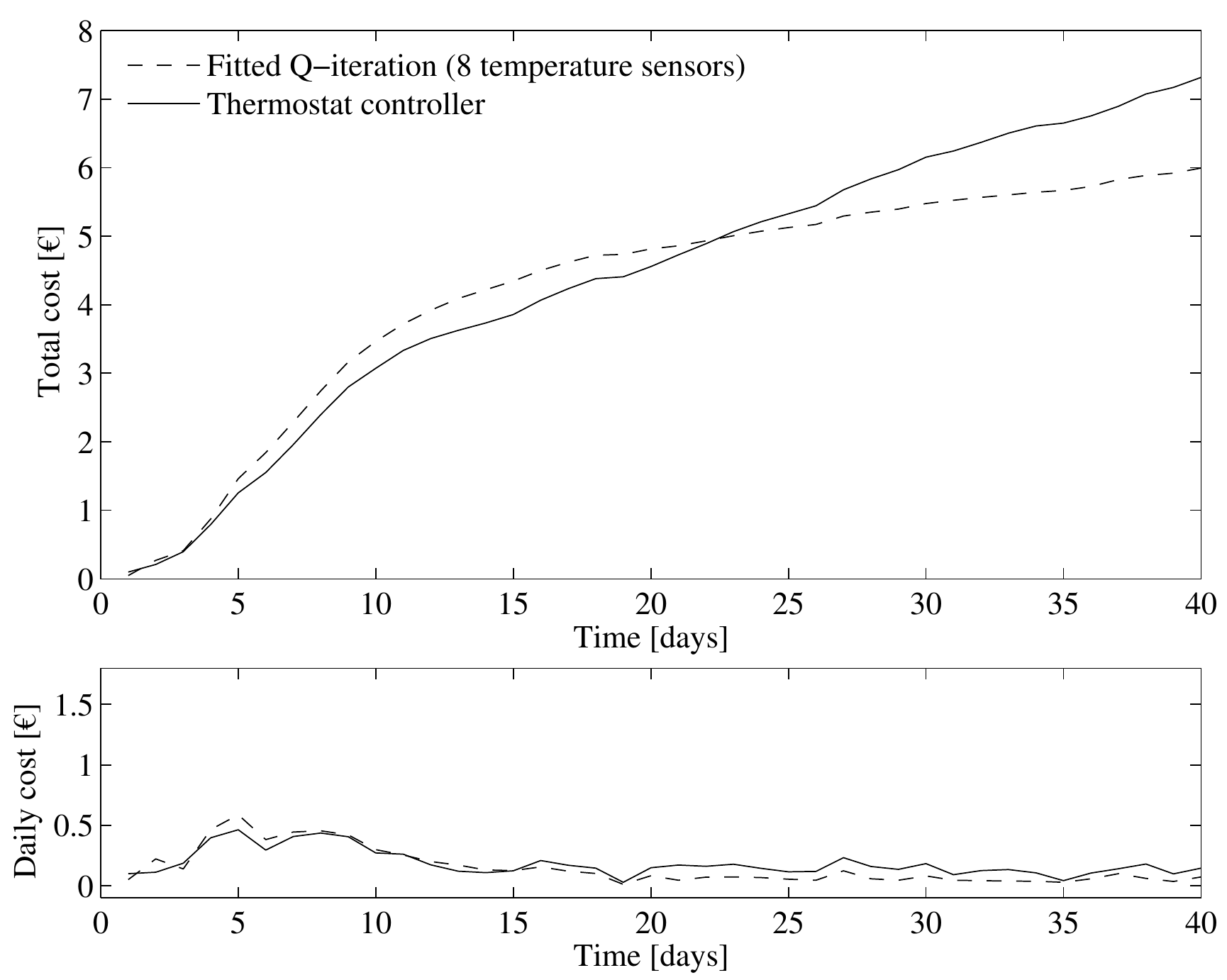}}
\put(-70,195){\textbf{a}}
\put(-75,62){\textbf{b}}
\end{picture}
\caption{Lab results of the learning agent using fitted Q-iteration (dashed line) and of the default thermostat (solid line) during 40 days. \textbf{a}, Cumulative energy cost. \textbf{b}, Daily energy cost.}
\label{boiler_lab_performance}
\end{figure}

Fig.~\ref{boiler_lab_performance} depicts the experimental results, spanning 40 days, of fitted Q-iteration and the default thermostat controller. 
The top plot of this figure indicates the cumulative costs of energy consumption and the bottom plot indicates the daily costs of energy consumption.
After 40 days, fitted Q-iteration was able to reduce the cost of energy consumption by \lab$\%$ compared to the default thermostat controller.
Furthermore, by excluding the first ten exploration days, this reduction increases to  28$\%$.

%
%\begin{figure}[t]
%\centerline{\includegraphics[width=9cm]{powers_lab_imb}}
%\caption{Experimental results using fitted Q-iteration combined with full state information. The top plot shows the temperature measurements. The middle plot indicates the hot tap water demand and the bottom plot indicates the price profile and power profile.}
%\label{snap_shot_lab_imbalance}
%\end{figure}

%It is important to note that the hot water demand is a unobservable stochastic variable. 

%\input{section/pol_repair}

\section{Conclusions and future work}
\label{conclusion}
This paper has demonstrated how an auto-encoder network can be used in combination with a well-established batch reinforcement learning algorithm, called fitted Q-iteration, to reduce the cost of energy consumption of an electric water heater.
The auto-encoder network was used to find a compact representation of the state vector.
In a series of simulation-based experiments using an electric water heater with 50 temperature sensors, the proposed method was able to converge to good policies much faster than when using the full state information. 
%The results indicated that this approach was able to reduce the learning period by learning a control policy in a compact representation.
%Alow-dimensional feature vectors can help mitigate the  and reduce  the learning period of the learning algorithm.
%This approach has been evaluated for an electric water heater, using realistic hot water profiles and electricity prices and using a non-linear simulation model of an electric water heater.
Compared to a default thermostat controller, the presented approach has reduced the cost of energy consumption by \dayahead$\%$ using day-ahead prices and by \balancing$\%$ using  imbalance prices. 

In a lab experiment,  fitted Q-iteration  has been  successfully applied  to an electric water heater with eight temperature sensors.
A reduction of the state vector did not improve the performance of fitted Q-iteration.
Compared to the thermostat controller, fitted Q-iteration was able to reduce the total cost of energy consumption by \lab$\%$ within 40 days of operation.

Based on the results of both experiments the following four conclusions can be drawn: 
(1) learning in a compact feature space can improve the quality of the control policy when the number of observations is relatively small (25 days);
(2) when the number of observations increases it is advisable to switch to higher state-space representation;
(3) when only a limited number of temperature sensors is available, i.e. 1-10 sensors, it is recommended to use the full state vector;
(4) when applied to a real-world stetting, fitted Q-iteration was able to obtain good control policies within 20 days of operation.

In our future research, we aim at developing a  method for selecting an appropriate state representation during the learning process.
A promising route is to construct experts, where each expert combines a learning algorithm with a different feature representation.
A metric based on the performance of each expert, as presented in~\cite{fonteneau2013batch}, could then be used to select the expert with the highest metric as described in~\cite{expert_advice}.

%\begin{figure}[t]
%\centerline{\includegraphics[width=9cm]{powers_lab_belp}}
%\caption{Experimental results using fitted Q-iteration combined with full state information. The top plot shows the temperature measurements. The middle plot indicates the hot tap water demand and the bottom plot indicates the price profile and power profile.}
%\label{boiler_lab_snapshot}
%\end{figure} 

%\begin{figure}[t]
%\centerline{\includegraphics[width=9cm]{cumulative_cost_lab}}
%\caption{Simulation results using fitted Q-iteration combined with full state information (eight temperature 
%sensors) and the default thermostat controller.}
%\label{boiler_lab_performance}
%\end{figure}

\ifapp
\appendices
\section*{Appendix A: Stratified tank model}
%\subsubsection{}
The simulation parameters of the stratified tank model are chosen to correspond with the real-world electric water heater (Table~\ref{parameters}).
The tank model has a buffer height $L_{\mathrm{b}}$,  diameter $d$, and water content of $V_{\mathrm{b}}$.  
Hot water can exit the buffer tank at the top and is replenished by cold water at the bottom of the buffer.
The water in the buffer tank is divided into $n_{\mathrm{d}}$ cylindrical, discs.
Each disc has a  uniform temperature $T_{i}$, thickness $L_{\mathrm{d},i}$, outer surface $A_{i}$ and conduction surface area $S_{i}$.
At each time step $t_{\mathrm{sim}}$ the temperature of all discs are updated as follows.
The discs can interact thermally through conduction and mixing.
The thermal losses of each disc are given by:
\begin{equation}
Q_{\mathrm{l},i} = A_{i} U_{i} (T_{\mathrm{a}}-T_{i}),
\label{losses}
\end{equation}
with  $U_{i}$ the heat loss coefficient of the surface and $T_{\mathrm{a}}$ the ambient room temperature.
The conduction effects are given by
\begin{align}
Q_{\mathrm{c},i} = k_{i} S_{i} \frac{(T_{i}-T_{i+1})}{L_{\mathrm{d},i}} + k_{i-1} S_{i-1} \frac{(T_{i}-T_{i-1})}{L_{\mathrm{d},i}},
\label{conduction}
\end{align}
where ${i+1}$ indicates the layer below layer $i$, layer ${i-1}$ indicates the layer above layer $i$, and $k$ is the thermal conductivity of water. 
The mixing effects caused by the tap demand are calculated as follows
\begin{align}
Q_{\mathrm{m},i} = \dot{m}_{i} C_{i} (T_{i}-T_{i+1}) + \dot{m}_{i-1}C_{i}(T_{i}-T_{i-1}),
\label{mix}
\end{align}
with $ \dot{m}_{i}$ the flow rate of the tap water demand, and $C_{i}$ the specific heat of water. 
An electric resistance can add heat to several discs
\begin{equation}
Q_{\mathrm{h},i} = \frac{P^{\mathrm{e}}}{n_{\mathrm{h}}},
\end{equation}
with $P^{\mathrm{e}}$ the electric heating power rating, and $n_{\mathrm{h}}$ the number discs that cover the heating resistance.
The mixing and turbulence effects caused by the heating element is modeled using the following heuristic. 
As long as an unstable layer is detected, i.e. a layer with a temperature  higher than the temperature of the layer above, an iteration is performed where the new temperature of the unstable layer is set to be equal to the average temperature of the layer above and below.
The iteration process continues until convergence is reached.
%\begin{figure}[t]
%\centerline{\includegraphics[width=\columnwidth]{expert_imbalance}}
%\caption{Two dimensional representation of the control policies. The left column of plots depicts the original control policies obtained with fitted Q-iteration. The right column of plots depicts the corresponding adjusted policies.}
%\label{expert_imbalance}
%\end{figure}

%
%\subsection*{B. Calculation of the state of charge}
%The state of charge of the buffer tank is defined by the ratio of the energy content relative to amount of useful thermal energy ($T_{j}\geq \Tmin$) and the maximum amount of useful thermal energy.
%\begin{equation}
%\soc =  \frac{\sum_{j=1}^{n_{\mathrm{d}}} V_{\mathrm{l},j}(T_{j}-\Tmin)}{\sum_{i=1}^{n_{\mathrm{d}}} V_{\mathrm{l},i}(\Tmax-\Tmin)} ,
%\end{equation}
%where $V_{\mathrm{l},j}$ indicates the volume of the $j^{\mathrm{th}}$ layer.
%The minimum state of charge $\socmin$ is set to $30\%$ and act as a buffer to guarantee the comfort of the end users. 
\begin{table} [h!]%
\caption{Simulation parameters of the electric water heater}
 \label{parameters}
\begin{center}
\begin{tabular}{ l l c |  l  l c}
\toprule
 Parameter& Value  & Unit & Parameter& Value  & Unit\\
 \midrule
	$A$       &  0.0393                &  [$\mathrm{m}^{2}$]& 	$S$        &    0.1963     & [$\mathrm{m}^{2}$]\\
	$C$      &   4185.5                & [J/(kgK)]& $P_{\mathrm{e}}$ &  2360     & [W]\\
	$n_{\mathrm{h}}$ &  5     & [-]& $U$                        &  0.8     & [W/$(\mathrm{m}^{2}$K)]\\   
  $L_{\mathrm{d}}$        &  0.025                & [m]&  $\Tamb$                  &  20     &[$^{\circ}$C]\\
	$k$       &  0.5944               & [W/(mK)]&  $t_{\mathrm{sim}}$                &  6     & [s]\\
	$n_{\mathrm{d}}$ &  50   & [-]&  $\Twater$                 &  10     & [$^{\circ}$C]\\
	$L_{\mathrm{b}}$ &  1.2   & [m]&  $d_{\mathrm{b}}$          &  0.5     & [m]\\
	$V_{\mathrm{l}}$ &  200   & [l]&  -         & -    & -\\

 \bottomrule
\end{tabular}
\end{center}      
\end{table}

\else
\fi

\ifack
\section*{Acknowledgment}
The authors would like to thank Davy Geysen, Geert Jacobs, Koen Vanthournout, and Jef Verbeeck  from Vito for providing us with the  lab setup.
This work was supported by a Ph.D. grant of the Institute for the Promotion of Innovation through Science and Technology in Flanders (IWT-Vlaanderen) and by Stable MultI-agent LEarnIng for neTworks (SMILE-IT).
\fi

\ifCLASSOPTIONcaptionsoff
  \newpage
\fi

\bibliographystyle{IEEEtran}  
\bibliography{references} 

\begin{thebibliography}{10}
\providecommand{\url}[1]{#1}
\csname url@samestyle\endcsname
\providecommand{\newblock}{\relax}
\providecommand{\bibinfo}[2]{#2}
\providecommand{\BIBentrySTDinterwordspacing}{\spaceskip=0pt\relax}
\providecommand{\BIBentryALTinterwordstretchfactor}{4}
\providecommand{\BIBentryALTinterwordspacing}{\spaceskip=\fontdimen2\font plus
\BIBentryALTinterwordstretchfactor\fontdimen3\font minus
  \fontdimen4\font\relax}
\providecommand{\BIBforeignlanguage}[2]{{%
\expandafter\ifx\csname l@#1\endcsname\relax
\typeout{** WARNING: IEEEtran.bst: No hyphenation pattern has been}%
\typeout{** loaded for the language `#1'. Using the pattern for}%
\typeout{** the default language instead.}%
\else
\language=\csname l@#1\endcsname
\fi
#2}}
\providecommand{\BIBdecl}{\relax}
\BIBdecl

\bibitem{outlook2013}
F.~Birol \emph{et~al.}, ``{World Energy Outlook 2013: Renewable Energy Outlook,
  An annual report released by the International Energy Agency},''
  \url{http://www.worldenergyoutlook.org/media/weowebsite/2013}, Paris, France,
  [Online: accessed July 21, 2015].

\bibitem{hastings1980ten}
B.~Hastings, ``Ten years of operating experience with a remote controlled water
  heater load management system at detroit edison,'' \emph{IEEE Trans. on Power
  Apparatus and Syst.}, no.~4, pp. 1437--1441, 1980.

\bibitem{SMARTBOILER}
K.~Vanthournout, R.~D'hulst, D.~Geysen, and G.~Jacobs, ``A smart domestic hot
  water buffer,'' \emph{IEEE Trans. on Smart Grid}, vol.~3, no.~4, pp.
  2121--2127, Dec. 2012.

\bibitem{energygov}
{U.S. Department of Energy}, ``Energy cost calculator for electric and gas
  water heaters,''
  \url{http://energy.gov/eere/femp/energy-cost-calculator-electric-and-gas-water-heaters-0{\#}output},
  [Online: accessed November 10, 2015].

\bibitem{diao2012electric}
R.~Diao, S.~Lu, M.~Elizondo, E.~Mayhorn, Y.~Zhang, and N.~Samaan, ``Electric
  water heater modeling and control strategies for demand response,'' in
  \emph{Proc. 2012 IEEE Power and Energy Society General Meeting,}, pp. 1--8.

\bibitem{Sandero}
S.~Iacovella, K.~Lemkens, F.~Geth, P.~Vingerhoets, G.~Deconinck, R.~D'Hulst,
  and K.~Vanthournout, ``Distributed voltage control mechanism in low-voltage
  distribution grid field test,'' in \emph{Proc. 4th IEEE PES Innov. Smart Grid
  Technol. Conf. (ISGT Europe)}, Oct 2013, pp. 1--5.

\bibitem{koch2011modeling}
S.~Koch, J.~L. Mathieu, and D.~S. Callaway, ``Modeling and control of
  aggregated heterogeneous thermostatically controlled loads for ancillary
  services,'' in \emph{Proc. 17th IEEE Power Sys. Comput. Conf. (PSCC)},
  Stockholm, Sweden, Aug. 2011, pp. 1--7.

\bibitem{Mathieu}
J.~Mathieu and D.~Callaway, ``State estimation and control of heterogeneous
  thermostatically controlled loads for load following,'' in \emph{Proc. 45th
  Hawaii Int. Conf. on System Science (HICSS)}, Maui, HI, Jan. 2012, pp.
  2002--2011.

\bibitem{sossan2013scheduling}
F.~Sossan, A.~M. Kosek, S.~Martinenas, M.~Marinelli, and H.~Bindner,
  ``Scheduling of domestic water heater power demand for maximizing {PV}
  self-consumption using model predictive control,'' in \emph{Proc. 4th IEEE
  PES Innov. Smart Grid Technol. Conf. (ISGT Europe)}, Oct 2013, pp. 1--5.

\bibitem{camacho2004model}
E.~F. Camacho and C.~Bordons, \emph{Model Predictive Control}, 2nd~ed.\hskip
  1em plus 0.5em minus 0.4em\relax London, UK: Springer London, 2004.

\bibitem{challengesMPC}
J.~Cigler, D.~Gyalistras, J.~{\v{S}}irok{\`y}, V.~Tiet, and L.~Ferkl, ``Beyond
  theory: the challenge of implementing model predictive control in
  buildings,'' in \emph{Proc. 11th {REHVA} World Congress ({CLIMA})}, Czech
  Republic, Prague, 2013.

\bibitem{ma2012model}
Y.~Ma, ``Model predictive control for energy efficient buildings,'' Ph.D.
  dissertation, University of California Berkeley, Mechanical Engineering,
  Berkeley, CA, 2012.

\bibitem{Maasoumy}
M.~Maasoumy, M.~Razmara, M.~Shahbakhti, and A.~Sangiovanni~Vincentelli,
  ``Selecting building predictive control based on model uncertainty,'' in
  \emph{Proc. American Control Conference (ACC)}, Portland, OR, June 2014, pp.
  404--411.

\bibitem{sutton1998reinforcement}
R.~S. Sutton and A.~G. Barto, \emph{Reinforcement Learning: An
  Introduction}.\hskip 1em plus 0.5em minus 0.4em\relax Cambridge, MA: {MIT
  Press}, 1998.

\bibitem{ernst2009reinforcement}
D.~Ernst, M.~Glavic, F.~Capitanescu, and L.~Wehenkel, ``Reinforcement learning
  versus model predictive control: a comparison on a power system problem,''
  \emph{IEEE Trans. Syst., Man, Cybern., Syst.}, vol.~39, no.~2, pp. 517--529,
  2009.

\bibitem{lange2010deep}
S.~Lange and M.~Riedmiller, ``Deep auto-encoder neural networks in
  reinforcement learning,'' in \emph{Proc. IEEE 2010 Int. Joint Conf. on Neural
  Networks (IJCNN)}, Barcelona, Spain, July 2010, pp. 1--8.

\bibitem{kara2012using}
E.~C. Kara, M.~Berges, B.~Krogh, and S.~Kar, ``Using smart devices for
  system-level management and control in the smart grid: A reinforcement
  learning framework,'' in \emph{Proc. 3rd IEEE Int. Conf. on Smart Grid
  Commun. (SmartGridComm)}, Tainan, Taiwan, Nov. 2012, pp. 85--90.

\bibitem{henze2003evaluation}
G.~P. Henze and J.~Schoenmann, ``Evaluation of reinforcement learning control
  for thermal energy storage systems,'' \emph{HVAC\&R Research}, vol.~9, no.~3,
  pp. 259--275, 2003.

\bibitem{WenZhen}
Z.~Wen, D.~O'Neill, and H.~Maei, ``Optimal demand response using device-based
  reinforcement learning,'' \emph{IEEE Trans. on Smart Grid}, vol.~6, no.~5,
  pp. 2312--2324, Sept 2015.

\bibitem{MarinaPscc2014}
M.~Gonz{\'a}lez, R.~Luis~Briones, and G.~Andersson, ``Optimal bidding of
  plug-in electric vehicles in a market-based control setup,'' in \emph{Proc.
  18th IEEE Power Sys. Comput. Conf. (PSCC)}, Wroclaw, Poland, 2014, pp. 1--7.

\bibitem{ernst2005tree}
D.~Ernst, P.~Geurts, and L.~Wehenkel, ``Tree-based batch mode reinforcement
  learning,'' \emph{Journal of Machine Learning Research}, pp. 503--556, 2005.

\bibitem{riedmiller2005neural}
M.~Riedmiller, ``Neural fitted {Q}-iteration--first experiences with a data
  efficient neural reinforcement learning method,'' in \emph{Proc. 16th
  European Conference on Machine Learning (ECML)}, vol. 3720.\hskip 1em plus
  0.5em minus 0.4em\relax Porto, Portugal: Springer, Oct. 2005, p. 317.

\bibitem{deepmind}
V.~Mnih, K.~Kavukcuoglu, D.~Silver, A.~A. Rusu, J.~Veness, M.~G. Bellemare,
  A.~Graves, M.~Riedmiller, A.~K. Fidjeland, G.~Ostrovski \emph{et~al.},
  ``Human-level control through deep reinforcement learning,'' \emph{Nature},
  vol. 518, no. 7540, pp. 529--533, 2015.

\bibitem{riedmiller2009reinforcement}
M.~Riedmiller, T.~Gabel, R.~Hafner, and S.~Lange, ``Reinforcement learning for
  robot soccer,'' \emph{Autonomous Robots}, vol.~27, no.~1, pp. 55--73, 2009.

\bibitem{fonteneau2008variable}
R.~Fonteneau, L.~Wehenkel, and D.~Ernst, ``Variable selection for dynamic
  treatment regimes: a reinforcement learning approach,'' in \emph{Proc.
  European Workshop on Reinforcement Learning (EWRL)}, Villeneuve d'Ascq,
  France, 2008.

\bibitem{adam2012experience}
S.~Adam, L.~Busoniu, and R.~Babu{\v{s}}ka, ``Experience replay for real-time
  reinforcement learning control,'' \emph{IEEE Trans. on Syst., Man, and
  Cybern., Part C: Applications and Reviews}, vol.~42, no.~2, pp. 201--212,
  2012.

\bibitem{RuelensBRLCluster}
F.~Ruelens, B.~Claessens, S.~Vandael, S.~Iacovella, P.~Vingerhoets, and
  R.~Belmans, ``Demand response of a heterogeneous cluster of electric water
  heaters using batch reinforcement learning,'' in \emph{Proc. 18th IEEE Power
  Sys. Comput. Conf. (PSCC)}, Wroc\l{}aw, Poland, Aug. 2014, pp. 1--8.

\bibitem{RuelensBRLDevice}
F.~{Ruelens}, B.~{Claessens}, S.~{Vandael}, B.~{De Schutter}, R.~{Babu\v{s}ka},
  and R.~{Belmans}, ``{Residential Demand Response Applications Using Batch
  Reinforcement Learning},'' \emph{Submitted to IEEE Trans. on Smart Grid
  (http://arxiv.org/pdf/1504.02125.pdf)}, Apr. 2015.

\bibitem{PILCO}
M.~Deisenroth and C.~E. Rasmussen, ``{PILCO}: A model-based and data-efficient
  approach to policy search,'' in \emph{Proceedings of the 28th International
  Conference on machine learning (ICML-11)}, 2011, pp. 465--472.

\bibitem{MABRL}
T.~Lampe and M.~Riedmiller, ``Approximate model-assisted neural fitted
  q-iteration,'' in \emph{2014 International Joint Conference on Neural
  Networks (IJCNN)}, July 2014, pp. 2698--2704.

\bibitem{GiuArxive}
\BIBentryALTinterwordspacing
G.~T. Costanzo, S.~Iacovella, F.~Ruelens, T.~Leurs, and B.~Claessens,
  ``Experimental analysis of data-driven control for a building heating
  system,'' \emph{CoRR}, vol. abs/1507.03638, 2015. [Online]. Available:
  \url{http://arxiv.org/abs/1507.03638}
\BIBentrySTDinterwordspacing

\bibitem{BellmanDP}
R.~Bellman, \emph{Dynamic Programming}.\hskip 1em plus 0.5em minus 0.4em\relax
  New York, NY: Dover Publications, 1957.

\bibitem{tim_brys}
W.~Curran, T.~Brys, M.~Taylor, and W.~Smart, ``Using {PCA} to efficiently
  represent state spaces,'' in \emph{The 12th European Workshop on
  Reinforcement Learning (EWRL 2015)}, Lille, France, 2015.

\bibitem{bertsekas1996neuro}
D.~Bertsekas and J.~Tsitsiklis, \emph{Neuro-Dynamic Programming}.\hskip 1em
  plus 0.5em minus 0.4em\relax Nashua, NH: Athena Scientific, 1996.

\bibitem{Scholz}
M.~Scholz and R.~Vig{\'a}rio, ``Nonlinear {PCA}: a new hierarchical approach.''
  in \emph{ESANN}, 2002, pp. 439--444.

\bibitem{kaelbling1996reinforcement}
L.~P. Kaelbling, M.~L. Littman, and A.~W. Moore, ``Reinforcement learning: A
  survey,'' \emph{Journal of Artificial Intelligence Research}, pp. 237--285,
  1996.

\bibitem{jordan2001realistic}
U.~Jordan and K.~Vajen, ``Realistic domestic hot-water profiles in different
  time scales: Report for the international energy agency, solar heating and
  cooling task ({IEA-SHC}),'' Universit{\"a}t Marburg, Marburg, Germany, Tech.
  Rep., 2001.

\bibitem{belpex}
``Belpex - {B}elgian power exchange,'' \url{http://www.belpex.be/}, [Online:
  accessed March 21, 2015].

\bibitem{elia}
``Elia - {B}elgian power exchange,'' \url{http://www.belpex.be/}, [Online:
  accessed March 21, 2015].

\bibitem{Dupont}
B.~Dupont, P.~Vingerhoets, P.~Tant, K.~Vanthournout, W.~Cardinaels,
  T.~De~Rybel, E.~Peeters, and R.~Belmans, ``{LINEAR} breakthrough project:
  Large-scale implementation of smart grid technologies in distribution
  grids,'' in \emph{Proc. 3rd IEEE PES Innov. Smart Grid Technol. Conf. (ISGT
  Europe)}, Berlin, Germany, Oct. 2012, pp. 1--8.

\bibitem{Scikit}
F.~Pedregosa, G.~Varoquaux, A.~Gramfort, V.~Michel, B.~Thirion, O.~Grisel,
  M.~Blondel, P.~Prettenhofer, R.~Weiss, V.~Dubourg \emph{et~al.},
  ``{Scikit-learn}: Machine learning in {Python},'' \emph{The Journal of
  Machine Learning Research}, vol.~12, pp. 2825--2830, 2011.

\bibitem{fonteneau2013batch}
R.~Fonteneau, S.~A. Murphy, L.~Wehenkel, and D.~Ernst, ``Batch mode
  reinforcement learning based on the synthesis of artificial trajectories,''
  \emph{Annals of Operations Research}, vol. 208, no.~1, pp. 383--416, 2013.

\bibitem{expert_advice}
M.~Devaine, P.~Gaillard, Y.~Goude, and G.~Stoltz, ``Forecasting electricity
  consumption by aggregating specialized experts,'' \emph{Machine Learning},
  vol.~90, no.~2, pp. 231--260, 2013.

\end{thebibliography}

\end{document}